%% file: neurips_2025.tex
\documentclass{article}

\usepackage[final]{neurips_2025}

\usepackage[utf8]{inputenc} % allow utf-8 input
\usepackage[T1]{fontenc}    % use 8-bit T1 fonts
\usepackage{hyperref}       % hyperlinks
\usepackage{url}            % simple URL typesetting
\usepackage{booktabs}       % professional-quality tables
\usepackage{amsfonts}       % blackboard math symbols
\usepackage{nicefrac}       % compact symbols for 1/2, etc.
\usepackage{microtype}      % microtypography
\usepackage{xcolor}         % colors
\usepackage{amsmath} 
\usepackage[table]{xcolor}
\usepackage{caption}
\usepackage{subcaption}
\usepackage{multirow}
\usepackage{makecell}
\usepackage{graphicx}

\title{CoViT: Instance-Correspondence Contrastive Learning for Vision Transformer}

\author{Yisen Wang \quad Zhirong Wu \quad Limin Wang \\ State Key Laboratory for Novel Software Technology \\ Nanjing University, China \\ \texttt{yswang@smail.nju.edu.cn}} 
\begin{document}\maketitle

\begin{abstract}
  Vision Transformers (ViT) excel in semantic understanding but fail to discriminate between object instances (e.g., identical embeddings for two dogs), limiting their use in instance-level tasks such as object detection and instance segmentation. We propose \textbf{Co}ntrastive \textbf{Vi}sion \textbf{T}ransformer (CoViT), a self-supervised learning framework that injects instance-awareness into ViT through geometry-guided contrastive learning. CoViT uniquely coordinates ViT’s attention maps and embeddings by constructing triplets: (1) Attention-guided masking: Refine multi-head attention via adaptive thresholding and morphological operations to generate instance masks, identifying foreground anchors; (2) Hardest contrastive mining: For each anchor, computing pairwise embedding similarities to select the intra-instance hardest positive (least similar patch within its mask) and inter-instance hardest negative (most similar patch from other instances), with intra-instance regions masked during negative search. These triplets drive a contrastive loss that simultaneously compresses intra-instance variance and expands inter-instance margins, forcing ViT to discern subtle geometric and appearance differences between instances. CoViT consistently achieves stable performance gains of over 2 AP points across multiple instance-level perception tasks by using ViT as backbone architecture. Notably, CoViT requires no extra decoders or labels, demonstrating that a pure ViT can learn instance-aware representations via inherent attention priors and targeted contrastive constraints. Code and models will be released.

\end{abstract}

\input{Illustrations/figure}
\input{Illustrations/table}

\input{chapters/1_intro}
\input{chapters/2_relate}
\input{chapters/3_method}

\input{chapters/4_experi}
\input{chapters/5_concl}

\bibliographystyle{unsrt}
\bibliography{egbib}

\end{document}

%% file: Illustrations/figure.tex
\newcommand{\pipeline}{
\begin{figure}[t]
    \vspace{-3mm}
    \centering
    \includegraphics[width=\linewidth]{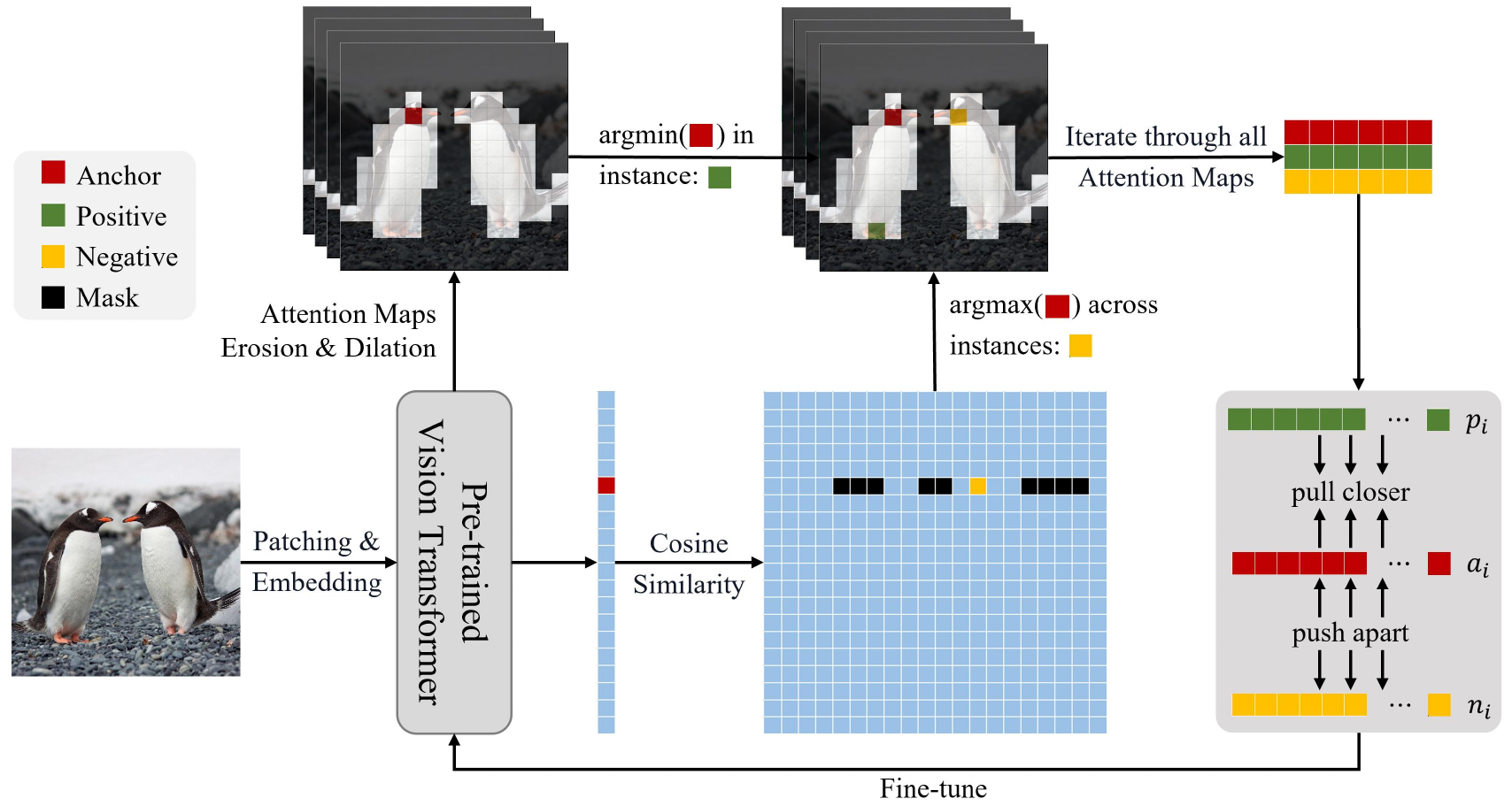}
    \caption{Overview of the proposed method. CoViT enhances ViT's instance discrimination by contrastive learning with attention-guided triplet mining. We extract refined instance regions from ViT's attention maps, then construct $\langle$\textbf{anchor}, \textbf{intra-instance dissimilar}, \textbf{inter-instance similar}$\rangle$ triplets using embedding similarities. The model is trained to reconcile intra-instance variance while amplifying inter-instance distinctions, enabling instance-aware representations without semantic loss.}
    \label{fig:pipeline}
    \vspace{-5mm}
\end{figure}
}

\newcommand{\attnmap}{
\begin{figure}[t]
    \vspace{-3mm}
    \centering
    \includegraphics[width=\linewidth]{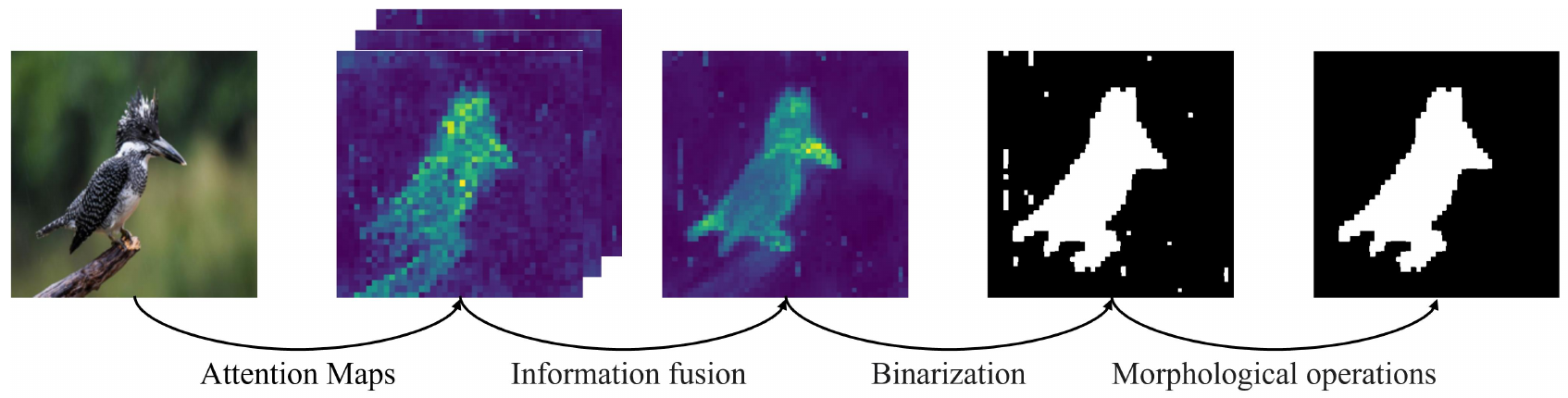}
    \caption{Illustration of the attention-map processing pipeline. The morphological stage applies dilation followed by erosion (1D1E) to connect fragmented foreground responses and then retract the expanded boundaries.}
    \label{fig:attnmap}
    \vspace{-5mm}
\end{figure}
}

%% file: Illustrations/table.tex
\newcommand{\cocotable}{
\begin{table}[!ht]
\centering
\small
\setlength{\tabcolsep}{3pt}
\vspace{-5mm}
\caption{Comparison of different methods on COCO dataset.}
\label{tab:coco_comparison}
\begin{tabular}{l c cccccc c cccccc}
\toprule
\multirow{2}{*}{Method} && \multicolumn{6}{c}{Baseline} && \multicolumn{6}{c}{Ours} \\
\cmidrule{3-8} \cmidrule{10-15}
&& AP$^b$ & AP$^b_{50}$ & AP$^b_{75}$ & AP$^m$ & AP$^m_{50}$ & AP$^m_{75}$ && AP$^b$ & AP$^b_{50}$ & AP$^b_{75}$ & AP$^m$ & AP$^m_{50}$ & AP$^m_{75}$ \\
\midrule
ViT-T~\cite{ViT} && 35.5 & 58.1 & 37.8 & 33.5 & 54.9 & 35.1 && 37.2 & 59.9 & 40.5 & 34.0 & 56.6 & 36.0 \\
ViT-S~\cite{ViT} && 40.2 & 63.1 & 43.4 & 37.1 & 59.9 & 39.3 && 42.0 & 64.9 & 45.8 & 37.9 & 61.4 & 40.2 \\
ViT-B~\cite{ViT} && 42.9 & 65.7 & 46.8 & 39.4 & 62.6 & 42.0 && 44.8 & 67.2 & 49.1 & 40.3 & 64.1 & 43.4 \\
ViT-L~\cite{ViT} && 45.7 & 68.9 & 49.4 & 41.5 & 65.6 & 44.6 && 47.3 & 69.4 & 51.9 & 42.4 & 66.9 & 45.6 \\
\midrule
ViTDet-T~\cite{ViTDet} && 35.7 & 57.7 & 38.4 & 33.5 & 54.7 & 35.2 && 37.4 & 60.3 & 40.9 & 34.1 & 57.1 & 36.3 \\
ViTDet-S~\cite{ViTDet} && 40.6 & 63.3 & 43.5 & 37.1 & 60.0 & 38.8 && 41.5 & 63.9 & 45.4 & 37.1 & 60.6 & 39.5 \\
ViTDet-B~\cite{ViTDet} && 43.2 & 65.8 & 46.9 & 39.2 & 62.7 & 41.4 && 44.3 & 66.6 & 48.5 & 39.6 & 63.3 & 42.5 \\
ViTDet-L~\cite{ViTDet} && 46.2 & 69.2 & 50.3 & 41.4 & 65.8 & 44.1 && 48.1 & 70.5 & 52.8 & 43.0 & 67.5 & 46.6 \\
\midrule
ViT-Adapter-T~\cite{ViTAdapter} && 41.1 & 62.5 & 44.3 & 37.5 & 59.7 & 39.9 && 43.0 & 64.6 & 47.4 & 38.0 & 61.4 & 41.1 \\
ViT-Adapter-S~\cite{ViTAdapter} && 41.7 & 62.8 & 45.3 & 36.9 & 59.5 & 39.8 && 44.2 & 65.7 & 48.5 & 38.8 & 62.4 & 41.9 \\
ViT-Adapter-B~\cite{ViTAdapter} && 47.0 & 68.2 & 51.4 & 41.8 & 65.1 & 44.9 && 49.6 & 70.6 & 54.0 & 43.6 & 67.7 & 46.9 \\
ViT-Adapter-L~\cite{ViTAdapter} && 48.7 & 70.1 & 53.2 & 43.3 & 67.0 & 46.9 && 51.1 & 72.8 & 55.5 & 45.0 & 69.5 & 48.7 \\
\midrule
ViT-B/32-FRCNN~\cite{ViTFRCNN} && 34.5 & 53.4 & 36.8 & - & - & - && 35.7 & 55.2 & 38.4 & - & - & -\\
ViT-B/16-FRCNN~\cite{ViTFRCNN} && 37.8 & 57.4 & 40.1 & - & - & - && 39.4 & 58.9 & 41.6 & - & - & -\\
\bottomrule
\vspace{-6mm}
\end{tabular}
\end{table}
}

\newcommand{\hicotable}{
\begin{table}[!ht]
\centering
\small
\setlength{\tabcolsep}{3pt}
\vspace{-5mm}
\caption{Comparison of different methods on HICO-DET dataset.}
\label{tab:hico_comparison}
\begin{tabular}{l c ccc c ccc c ccc c ccc}
\toprule
\multirow{4}{*}{Method}  && \multicolumn{7}{c}{Baseline} && \multicolumn{7}{c}{Ours} \\
&& \multicolumn{3}{c}{Default} && \multicolumn{3}{c}{Known Object} && \multicolumn{3}{c}{Default} && \multicolumn{3}{c}{Known Object} \\
\cmidrule{3-5} \cmidrule{7-9} \cmidrule{11-13}  \cmidrule{15-17}
&& Full & Rare & \makecell{Non-\\Rare} && Full & Rare & \makecell{Non-\\Rare} && Full & Rare & \makecell{Non-\\Rare} && Full & Rare & \makecell{Non-\\Rare} \\
\toprule
DEFR~\cite{DEFR} && 32.35 & 33.45 & 32.02 && - & - & - && 34.89 & 36.03 & 34.48 && - & - & - \\
ViPLO-S~\cite{ViPLO} && 34.95 & 33.83 & 35.28 && 38.15 & 36.77 & 38.56 && 37.14 & 35.93 & 37.84 && 40.35 & 38.98 & 40.47\\
ViPLO-L~\cite{ViPLO} && 37.22 & 35.45 & 37.75 && 40.61 & 38.82 & 41.15 && 39.25 & 37.42 & 39.81 && 42.75 & 40.98 & 43.26 \\
\bottomrule
\vspace{-6mm}
\end{tabular}
\end{table}
}

\newcommand{\ablationtable}{
\begin{table}[!ht]
\centering
\small
\vspace{-5mm}
\caption{Ablation Study on the HICO-DET dataset.}

% 设置每列为 0.45\textwidth 以便留出间距
\begin{tabular}{cc}
% 第一行
\begin{subtable}[b]{0.48\textwidth}
\centering
\begin{tabular}{c|ccc}
\toprule
Attn Head & Full & Rare & Non-Rare \\
\midrule
Base   & 32.35 & 33.45 & 32.02 \\
Rand 1 & 33.72{\color{blue}$\uparrow_{1.37}$} & 34.88 & 33.41 \\
Emp. 4 & 34.67{\color{blue}$\uparrow_{2.32}$} & 35.77 & 34.23 \\ 
Rand 4 & 34.12{\color{blue}$\uparrow_{1.77}$} & 35.34 & 33.49 \\ \rowcolor{gray!15}
Average  & 34.89{\color{blue}$\uparrow_{2.54}$} & 36.03 & 34.48 \\
\bottomrule
\end{tabular}
\caption{Study on Attention Head Selection.}
\label{tab:ablation:attn_head}
\end{subtable}
&
\begin{subtable}[b]{0.48\textwidth}
\centering
\begin{tabular}{c|ccc}
\toprule
Method & Full & Rare & Non-Rare \\
\midrule
Base & 32.35 & 33.45 & 32.02 \\ %\rowcolor{gray!15}
\rowcolor{gray!15} 1D1E    & 34.89{\color{blue}$\uparrow_{2.54}$} & 36.03 & 34.48 \\ 
\rowcolor{white} 1E1D    & 34.02{\color{blue}$\uparrow_{1.67}$} & 35.14 & 33.59 \\
2D2E   & 34.57{\color{blue}$\uparrow_{2.22}$} & 35.78 & 34.14 \\
2E2D  & 33.84{\color{blue}$\uparrow_{1.49}$} & 34.96 & 33.53 \\ 
\bottomrule
\end{tabular} 
\caption{Study on Morphological Operations.}
\label{tab:ablation:erode_dilate}
\end{subtable}
\\

% 第二行
\begin{subtable}[b]{0.48\textwidth}
\centering
\begin{tabular}{c|ccc}
\toprule
Loss Type & Full & Rare & Non-Rare \\
\midrule \rowcolor{gray!15}
TL   & 34.89{\color{blue}$\uparrow_{2.54}$} & 36.03 & 34.48 \\ 
\rowcolor{white} NCL  & 33.87{\color{blue}$\uparrow_{1.52}$} & 35.11 & 33.53 \\ 
MML  & 34.22{\color{blue}$\uparrow_{1.87}$} & 35.34 & 33.86 \\
HEAL & 34.74{\color{blue}$\uparrow_{2.39}$} & 35.86 & 34.31 \\
MVCL & 34.24{\color{blue}$\uparrow_{1.89}$} & 35.46 & 33.91 \\
\bottomrule
\end{tabular}
\caption{Study on Contrastive Loss Type.}
\label{tab:ablation:loss_type}
\end{subtable}
&
\begin{subtable}[b]{0.48\textwidth}
\centering
\begin{tabular}{c|ccc}
\toprule
Method & Full & Rare & Non-Rare \\
\midrule
Base & 32.35 & 33.45 & 32.02 \\  
Base+M & 34.72{\color{blue}$\uparrow_{2.37}$} & 35.87 & 34.34 \\
DINO+M & 34.86{\color{blue}$\uparrow_{2.51}$} & 35.98 & 34.45 \\
CLIP+M  & 34.83{\color{blue}$\uparrow_{2.48}$} & 35.94 & 34.43 \\ \rowcolor{gray!15}
MAE+M & 34.89{\color{blue}$\uparrow_{2.54}$} & 36.03 & 34.48 \\
\bottomrule
\end{tabular}
\caption{Study on Pre-train Weight.}
\label{tab:ablation:weight_init}
\end{subtable}
\\

% 第三行
\begin{subtable}[b]{0.48\textwidth}
\centering
\begin{tabular}{c|ccc}
\toprule
Method & Full & Rare & Non-Rare \\
\midrule 
Base  & 32.35 & 33.45 & 32.02 \\ 
Method 1 & 32.97{\color{blue}$\uparrow_{0.62}$} & 34.11 & 32.63 \\ 
Method 2 & 34.32{\color{blue}$\uparrow_{1.97}$} & 35.44 & 33.96 \\ \rowcolor{gray!15}
Method 3 & 34.89{\color{blue}$\uparrow_{2.54}$} & 36.03 & 34.48 \\
\bottomrule
\end{tabular}
\caption{Study on Positive Sample Selection.}
\label{tab:ablation:triplet_type}
\end{subtable}
&
\begin{subtable}[b]{0.48\textwidth}
\centering
\begin{tabular}{c|ccc}
\toprule
Method & Full & Rare & Non-Rare \\
\midrule
Base-B/16 & 32.35 & 33.45 & 32.02 \\  
MAE-B/16 & 34.89{\color{blue}$\uparrow_{2.54}$} & 36.03 & 34.48 \\
MAE-L/16 & 37.56{\color{blue}$\uparrow_{5.21}$} & 39.58 & 36.25 \\ \rowcolor{gray!15}
MAE-H/14  & 39.83{\color{blue}$\uparrow_{7.48}$} & 42.24 & 38.23 \\ 
\bottomrule
\end{tabular}
\caption{Study on MAE Pre-train Weight.}
\label{tab:ablation:weight_again}
\end{subtable}
\vspace{-6mm}
\end{tabular}
\end{table}
}

%% file: chapters/1_intro.tex
\section{Introduction}  
Vision Transformers (ViT)~\cite{ViT} have emerged as cornerstones in modern computer vision, demonstrating exceptional capabilities in semantic understanding through pre-training paradigms such as DeiT~\cite{DeiT}, MAE~\cite{MAE}, and DINO~\cite{DINO}. These methods excel at tasks requiring holistic scene interpretation, including image classification and semantic segmentation, by learning representations that emphasize semantic coherence across image regions. However, a critical gap remains unaddressed. ViT inherently lacks the ability to distinguish between individual instances of the same semantic category. For example, when there are multiple different instances with the same semantics in the image that the ViT needs to process, such as two dogs, the embedding output of the ViT will produce roughly the same representation. This limitation fundamentally restricts their utility in instance-centric applications like instance segmentation, where precise discrimination between objects is paramount.  

The underlying issue stems from the semantic-driven nature of existing pre-training objectives. Methods like MAE prioritize reconstructing masked patches through global context aggregation, while DINO enforces consistency between augmented views of an image. Although effective for learning invariant semantic features, such strategies inadvertently suppress fine-grained instance-specific cues. Consequently, the self-attention mechanisms in ViTs tend to capture broad semantic relationships rather than delineating boundaries between adjacent instances. Even state-of-the-art models struggle to maintain intra-instance feature consistency while maximizing inter-instance disparity, leading to ambiguous embeddings that hinder downstream instance-level tasks.  

To bridge this gap, we introduce Contrastive Vision Transformer (CoViT), a contrastive learning framework designed to endow pre-trained ViTs with instance discrimination capabilities without architectural modifications. Our approach is grounded in two pivotal observations. First, thresholded and morphologically refined attention maps, despite their imperfections, provide spatial cues that approximate instance regions. Second, analyzing embedding similarities reveals systematic errors: embeddings of distinct instances often exhibit unexpectedly high similarity, while patches within the same instance may show undesirably low similarity. Leveraging these insights, CoViT constructs informative triplets by identifying intra-instance hard positives (regions within the same instance with minimal similarity to the anchor) and inter-instance hard negatives (regions from different instances with maximal similarity to the anchor). These triplets are derived through a synergistic analysis of attention maps and embedding similarities, enabling targeted contrastive optimization.  

The CoViT framework imposes dual objectives: reducing feature distance between anchors and their intra-instance hard positives to enhance instance cohesion, while increasing the margin between anchors and inter-instance hard negatives to mitigate cross-instance confusion. This strategy directly addresses the limitations of conventional contrastive learning, which typically operates on image-level or patch-level pairs without explicit instance guidance. Experiments demonstrate that CoViT significantly enhances the instance awareness of pre-trained ViTs. When applied to DINO-Base, our method improves mask AP on COCO instance segmentation by 3.2\% and boosts few-shot instance recognition accuracy on PASCAL VOC by 12.7\%, all while preserving the model’s original semantic representation strength.  

Our work makes three core contributions. First, we systematically identify and analyze the inability of semantic-focused ViTs to discriminate instances, linking this limitation to the objectives of mainstream pre-training paradigms. Second, we propose CoViT, a novel framework that jointly exploits attention maps and embedding similarities to mine instance-aware triplets for contrastive fine-tuning. Third, we empirically validate CoViT’s versatility across multiple instance-level tasks and ViT architectures, providing insights into the interplay between semantic and instance-level feature learning. By reconciling the semantic richness of pre-trained ViTs with instance-sensitive adaptation, this work opens new avenues for deploying ViTs in scenarios requiring precise instance discrimination.  

%% file: chapters/2_relate.tex
\section{Related Work}

The rapid evolution of vision transformers has fundamentally transformed semantic understanding in computer vision, yet the nuanced challenge of distinguishing between object instances remains inadequately addressed. Current self-supervised paradigms like MAE~\cite{MAE} achieve remarkable semantic reconstruction through masked patch prediction, but their pixel-level optimization inherently overlooks fine-grained instance boundaries. While subsequent variants like DINO~\cite{DINO, DINOv2} enhance feature uniformity by integrating multi-task objectives from iBOT~\cite{iBOT} and SwAV~\cite{SwAV}, and ViTDet~\cite{ViTDet} demonstrates the latent detection potential of standard ViT architectures through strategic multi-scale convolution, these methods remain fundamentally constrained by their semantic-first design philosophy. The core limitation persists: existing pre-training objectives align poorly with instance-level tasks requiring discrimination between semantically identical entities.

Contrastive learning approaches have attempted to bridge this gap through various strategies. Early frameworks like MoCo~\cite{MoCo, MoCov2, MoCov3} and SimCLR~\cite{SimCLR, SimCLRv2} leverage negative sampling to amplify feature discriminability, but their reliance on random or class-agnostic negatives often introduces noisy training signals. Non-contrastive alternatives like DINO avoid negative sampling through self-distillation, yet sacrifice local instance details in favor of global feature consistency. Recent innovations reveal promising directions: BOAT~\cite{BOAT} introduces dynamic token clustering in both spatial and feature domains to sharpen instance boundaries, while CRATE~\cite{CRATE} establishes mathematical interpretability for instance segmentation through sparse coding principles. Medical imaging studies like ViT-Med~\cite{ViT-Med} further demonstrate ViT's instance-aware potential, with models achieving diagnostic-grade accuracy through attention map interpretability. However, methods like MaskContrast~\cite{MaskContrast}—which directly utilize ViT attention maps for pseudo-mask generation—remain hampered by the inherent semantic biases of pre-trained attention mechanisms, particularly in complex scenes with occluded or overlapping instances.

Architectural modifications present another dimension of exploration, though often at the cost of generalizability. Vision Reservoir~\cite{ViR} and Vision Mamba~\cite{ViM} dramatically reduces parameters through reservoir computing layers but sacrifices spatial sensitivity crucial for instance localization. Domain-specific adaptations like CVFormer's ~\cite{CVFormer}circular-view attention for autonomous driving and hybrid CNN-ViT~\cite{CNNViT} designs in behavior recognition systems achieve task-specific gains through carefully engineered inductive biases. MultiMAE~\cite{MultiMAE} extends masked autoencoding to multimodal inputs, revealing that joint training with auxiliary modalities can enhance instance discrimination—a finding that underscores the untapped potential of leveraging ViT's native attention mechanisms rather than modifying core architecture. CaiT~\cite{CaiT} addresses the instability of deep ViT training by introducing the LayerScale mechanism and decoupling class-attention layers, achieving SOTA performance on ImageNet with fewer parameters. Its channel-wise residual scaling strategy offers new insights for future architectural designs. DETR~\cite{DETR} pioneers the application of Transformer in end-to-end object detection, eliminating the need for NMS and anchors through bipartite matching loss. Despite limitations in small object detection and training efficiency, its serialized prediction paradigm has inspired subsequent research in instance-aware tasks, such as Deformable DETR~\cite{DDETR}, Conditional DETR~\cite{CDETR}, DAB-DETR~\cite{DABDETR}, Group DETR~\cite{GroupDETR}, Co-DETR~\cite{CoDETR}, and others.
 
These collective insights motivate our approach to instance discrimination without architectural alteration. By developing entropy-guided morphological refinement of attention maps—inspired by medical image analysis techniques—we enable precise instance region extraction even in cluttered scenes. Coupled with adaptive triplet mining that contrasts intra-instance variance against inter-instance similarity, the framework dynamically identifies challenging cases where current representations falter. This dual strategy preserves ViT's semantic strengths while injecting instance awareness, bridging the critical gap between self-supervised pre-training objectives and instance-level task requirements across segmentation, tracking, and recognition domains.

%% file: chapters/3_method.tex
\section{Approach}

\subsection{Overview}   

Our method enhances ViT with instance discrimination by leveraging attention-guided contrastive learning. The framework operates in three interconnected stages, as illustrated in Figure~\ref{fig:pipeline}. First, we extract and refine the ViT’s multi-head attention maps to generate binary spatial masks that localize potential instance regions. This involves aggregating attention heads, adaptive thresholding, and morphological operations to suppress noise and delineate instance boundaries. Second, using these masks and the ViT’s patch embeddings, we mine challenging correspondence pairs: \textbf{intra-instance dissimilar points} (regions within the same instance that are semantically distinct in embedding space) and \textbf{inter-instance similar points} (regions across different instances with high embedding similarity). Finally, we design a contrastive objective that explicitly optimizes the ViT to minimize intra-instance variance while maximizing inter-instance separation. By iterating between attention-based instance localization and semantic correspondence learning, the model progressively learns to disentangle instances even in crowded scenes, without relying on manual instance-level annotations.  

\pipeline

\subsection{Binary Attention Map Generation}  

The core of our instance localization approach lies in transforming the self-attention patterns of ViT into spatially meaningful binary masks. While standard ViT compute multi-head attention to model global dependencies, we observe that the aggregated attention maps implicitly highlight regions corresponding to cohesive semantic units, such as object parts or entire instances. To leverage this property, we propose a hierarchical process to convert raw attention maps into instance-aware binary masks.

Given an input image divided into $ N = H \times W $ patches, let $ \mathbf{A} \in \mathbb{R}^{M \times N \times N} $ represent the attention maps from $ M $ self-attention heads. Directly using individual attention heads introduces noise due to their diverse focus—some heads emphasize textures, while others capture global structures. To mitigate this issue, we average across all heads to obtain a consolidated attention map:

\begin{equation}
    \mathbf{\bar{A}} = \frac{1}{M}\sum_{m=1}^M \mathbf{A}^{(m)}, \quad \mathbf{\bar{A}} \in \mathbb{R}^{N \times N},
\end{equation}

where $ \mathbf{\bar{A}}(i,j) $ represents the normalized attention score between query patch $ i $ and key patch $ j $. We then reshape $ \mathbf{\bar{A}} $ into a spatial tensor $ \mathbf{\hat{A}} \in \mathbb{R}^{N \times H \times W} $, where each query patch $ i $ corresponds to a 2D attention map $ \mathbf{\hat{A}}_i \in \mathbb{R}^{H \times W} $. This transformation aligns the attention scores with the original spatial layout of the patches.

Each spatial attention map $ \mathbf{\hat{A}}_i $ reflects how strongly patch $ i $ attends to other regions in the image. To distinguish foreground (instance regions) from background, we binarize $ \mathbf{\hat{A}}_i $ using a query-specific adaptive threshold:

\begin{equation}
    \mathbf{B}_i(h,w) = \begin{cases} 
1 & \text{if } \mathbf{\hat{A}}_i(h,w) \geq \tau \cdot \max(\mathbf{\hat{A}}_i), \\
0 & \text{otherwise},
\end{cases}
\end{equation}

where $ \tau \in (0,1) $ is a hyperparameter controlling sensitivity to weak attention signals. Unlike fixed thresholds, this adaptive strategy accounts for variations in attention magnitudes across different queries and images. For example, a patch centered on a salient object, such as a dog’s head, typically exhibits higher attention concentration, allowing for stricter thresholds ($ \tau \approx 0.7 $). Conversely, a patch on a homogeneous region, like the sky, may require a lower $ \tau $ to retain sparse activations.

The initial binary maps $ \mathbf{B}_i $ often contain fragmented foreground responses, small holes, and irregular boundaries due to the coarse granularity of ViT attention. To close short gaps while preserving the overall foreground support, we apply dilation followed by erosion (1D1E), corresponding to a morphological closing operation:

\textbf{Dilation}: Connect nearby foreground responses and fill small gaps using a $ 3 \times 3 $ kernel $ k_{\text{dilate}} $:
\begin{equation}
    \mathbf{B}_i^{\text{dilate}} = \mathbf{B}_i \oplus k_{\text{dilate}},
\end{equation}

where $ \oplus $ denotes the dilation operator.

\textbf{Erosion}: Retract the expanded boundaries while retaining the newly connected foreground structures:
\begin{equation}
    \mathbf{M}_i = \mathbf{B}_i^{\text{dilate}} \ominus k_{\text{erode}},
\end{equation}

where $ \ominus $ is the erosion operator applied with a $ 3 \times 3 $ kernel $ k_{\text{erode}} $.

The refined masks $ \mathbf{M}_i $ may still contain overlapping regions from multiple instances. To isolate individual instances, we extract connected components from $ \mathbf{M}_i $. Specifically, for each query patch $ i $, we treat its non-zero regions in $ \mathbf{M}_i $ as candidate instance areas and apply a connected-component labeling algorithm:

\begin{equation}
    \mathcal{S}_i = \text{ConnectedComponents}(\mathbf{M}_i),
\end{equation}

where $ \mathcal{S}_i $ is the set of disjoint regions (instances) associated with query $ i $. The final instance mask set $ \mathcal{S} = \bigcup_{i=1}^N \mathcal{S}_i $ aggregates candidates from all queries, ensuring comprehensive coverage of potential instances.

\attnmap

\subsection{Correspondence Mining}  
The goal of correspondence mining is to identify challenging pairs that expose the ViT’s limitations in distinguishing instances. Specifically, we focus on two types of correspondences: regions within the same instance that are semantically distinct in embedding space, referred to as intra-instance dissimilar points, and regions across different instances that exhibit high semantic similarity in embedding space, referred to as inter-instance similar points. By optimizing these pairs, we encourage the ViT to refine its embeddings to better capture fine-grained instance boundaries, ultimately improving its ability to distinguish between objects.

For each query patch \( i \), let \( S_i \in \mathcal{S} \) represent the instance region (connected component) in the binary attention mask \( \mathbf{M}_i \). A patch \( i \) is considered a valid anchor if it satisfies the following condition:

\begin{equation}
    |S_i| \geq \gamma,
\end{equation}

where \( |S_i| \) denotes the number of patches in \( S_i \), and \( \gamma \) is a minimum instance size threshold (e.g., \( \gamma = 5 \) patches). This condition ensures that noisy anchors arising from fragmented or undersized regions are excluded, allowing us to focus on meaningful regions for further analysis. With valid anchors identified, we proceed to mine both intra-instance dissimilarities and inter-instance similarities.

To uncover intra-instance dissimilarities, we search within the instance \( S_i \) for the patch \( j^* \) that is least similar to \( i \) in embedding space. Formally, this is expressed as:

\begin{equation}
    j^* = \underset{j \in S_i \setminus \{i\}}{\text{argmin}} \ \mathbf{S}(i,j),
\end{equation}

where \( \mathbf{S} \in [-1,1]^{N \times N} \) is the precomputed cosine similarity matrix of patch embeddings. Minimizing \( \mathbf{S}(i,j^*) \) highlights regions where the ViT struggles to maintain intra-instance consistency—such as a dog’s leg and head, which may have low semantic similarity despite belonging to the same instance. To avoid trivial solutions, such as selecting boundary artifacts, we constrain \( j^* \) to lie within the dilated mask \( S_i \oplus k_{\text{dilate}} \), ensuring geometric coherence while focusing on meaningful intra-instance relationships.

Next, to address inter-instance similarities, we identify the patch \( k^* \) from other instances that is most similar to \( i \). To achieve this, we first suppress similarities to patches within \( S_i \) by masking intra-instance regions. This is done by defining:

\begin{equation}
    \mathbf{\tilde{S}}(i,k) = \begin{cases} \mathbf{S}(i,k), & k \notin S_i, \\ -\infty, & k \in S_i, \end{cases}
\end{equation}

where assigning \( -\infty \) to intra-instance positions guarantees that they cannot be selected by the subsequent maximization. We then perform hard negative mining by selecting:

\begin{equation}
    k^* = \underset{k}{\text{argmax}} \ \mathbf{\tilde{S}}(i,k).
\end{equation}

Maximizing \( \mathbf{S}(i,k^*) \) focuses on inter-instance confusion points, such as two different dogs with similar appearances. Penalizing these pairs helps the model disentangle embeddings of distinct instances. To ensure \( k^* \) belongs to a different instance, we verify that \( k^* \notin S_j \) for any \( S_j \in \mathcal{S} \) overlapping with \( S_i \).

Finally, a triplet \( (i, j^*, k^*) \) is retained only if it satisfies specific criteria. First, the triplet must exhibit semantic hardness, defined as:

\begin{equation}
    \mathbf{S}(i,k^*) > \mathbf{S}(i,j^*) - \alpha,
\end{equation}

where \( \alpha > 0 \) is a margin threshold. This ensures that the triplet is sufficiently challenging and provides meaningful training signals. Additionally, both \( j^* \) and \( k^* \) must lie within the dilated and eroded masks of their respective instances to avoid edge artifacts. These constraints collectively ensure that the mined triplets are both semantically and geometrically meaningful, enabling the model to learn robust and fine-grained instance representations.

By systematically addressing intra-instance dissimilarities and inter-instance similarities, and carefully selecting challenging triplets, our approach forces the ViT to refine its embeddings. This refinement not only enhances the model’s ability to recognize subtle distinctions between objects but also improves its overall performance in tasks requiring precise instance localization and segmentation.

\subsection{Contrastive Learning}  
The core objective of our framework is to refine the ViT’s embeddings to achieve strong intra-instance cohesion and inter-instance separation. To accomplish this, we design a hybrid contrastive loss that explicitly optimizes the challenging triplets mined in §3.3. Unlike conventional contrastive learning paradigms, which rely on predefined positive and negative pairs (e.g., via data augmentation), our approach dynamically identifies semantically meaningful hard pairs guided by the ViT’s own attention patterns. This targeted refinement enhances the model’s ability to discriminate between instances while preserving intra-instance consistency.

For each valid triplet \((i, j^*, k^*)\), where \(i\) is the anchor, \(j^*\) represents the intra-instance dissimilar point, and \(k^*\) denotes the inter-instance similar point, we impose a margin-based constraint to restructure the embedding space:

\begin{equation}
    \mathcal{L}_{\text{tri}} = \frac{1}{|\mathcal{T}|} \sum_{(i,j^*,k^*) \in \mathcal{T}} \max\left( \mathbf{S}(i,k^*) - \mathbf{S}(i,j^*) + \alpha, 0 \right),
\end{equation}

where \(\alpha > 0\) is a margin hyperparameter, and \(\mathbf{S}\) represents the cosine similarity matrix. This loss ensures that the similarity between \(i\) and its intra-instance hard positive \(j^*\) exceeds that of the inter-instance hard negative \(k^*\) by at least \(\alpha\). By focusing on hard pairs, the model learns to bridge semantic gaps within the same instance—such as aligning a cat’s tail with its head—while amplifying distinctions between visually similar regions across different instances, like two adjacent birds.

To account for varying difficulty levels across triplets, we dynamically adjust \(\alpha\) based on the hardness of the pair:

\begin{equation}
    \alpha(i) = \alpha_{\text{base}} + \beta \cdot \left( \mathbf{S}(i,k^*) - \mathbf{S}(i,j^*) \right),
\end{equation}

where \(\beta\) controls the adaptation rate. This ensures that already well-separated pairs (\(\mathbf{S}(i,j^*) \gg \mathbf{S}(i,k^*)\)) contribute less to the loss, preventing over-optimization of trivial cases and allowing the model to focus on more challenging scenarios.

While the triplet loss addresses inter-instance confusion, we further enhance intra-instance uniformity by maximizing the similarity between the anchor \(i\) and all patches within its instance region \(S_i\). Inspired by the InfoNCE loss, we formulate this as:

\begin{equation}
    \mathcal{L}_{\text{intra}} = -\frac{1}{|\mathcal{T}|} \sum_{(i,j^*,k^*) \in \mathcal{T}} \frac{1}{|P_i|} \sum_{m \in P_i} \log \frac{\exp\left( \mathbf{S}(i,m) / t \right)}{\sum_{p \in P_i} \exp\left( \mathbf{S}(i,p) / t \right) + \exp\left( \mathbf{S}(i,k^*) / t \right)},
\end{equation}

where \(P_i = S_i \setminus \{i\}\) denotes all intra-instance positive patches and \(t > 0\) is a temperature parameter. Averaging over every \(m \in P_i\) prevents the loss from favoring only the single hardest positive, while including the mined inter-instance hard negative \(k^*\) in the denominator provides contrastive competition. The term therefore improves coherence across the full instance while repelling the most confusing external patch.

Traditional contrastive losses, such as those used in SimCLR, treat augmented views of the same image as positives and all other images as negatives. However, in our scenario, the relationships between positives and negatives are more nuanced. Positives are heterogeneous, as intra-instance regions may exhibit significant appearance variations. Negatives are structured, as inter-instance confusion is often spatially localized—for example, adjacent objects in an image. Our hybrid loss explicitly models these properties, capturing instance-level geometric relationships that generic losses fail to address. By combining triplet-based inter-instance separation with intra-instance cohesion, our framework achieves a more refined and semantically meaningful embedding space.

%% file: chapters/4_experi.tex
\section{Experiment}
% ViTDet
% ViT-Adapter
% ViT-FRCNN
% 
% CoDETR
% Group DETR
% EVA-02

% DEFR
% ViPLO
\subsection{Experimental Setup}  
We evaluate our framework on COCO 2017 (object detection/instance segmentation) and HICO-DET(HOI detection), using ViT-B/16 backbones pretrained with MAE, CLIP vision encoder, DINO, and supervised ImageNet-1K initialization. Baselines include state-of-the-art instance-aware methods under identical hardware configurations.  

For attention mask generation, we set adaptive threshold \( \tau = 0.65 \) with entropy-based adjustment: \( \tau(i) = 0.7 - 0.15 \cdot H(\mathbf{\hat{A}}_i)/\log N \), where \( H(\cdot) \) measures attention map entropy. Morphological operations use \( 3 \times 3 \) kernels, filtering instances smaller than \( \gamma = 5 \) patches. Correspondence mining employs margin \( \alpha_{\text{base}} = 0.2 \), adaptation rate \( \beta = 0.1 \), and intra-instance temperature \( t = 0.1 \). Hybrid contrastive losses are weighted as \( \lambda_{\text{tri}} = 0.4 \), \( \lambda_{\text{intra}} = 0.6 \), optimized via grid search on a 5\% COCO subset.  

Training uses AdamW (\( \text{lr} = 1\text{e-4} \), weight decay 0.05) with 1024×1024 inputs and 64 batch size. We implement a three-phase protocol: (1) 10-epoch warmup with frozen ViT, training only morphological and projection layers; (2) Gradual ViT unfreezing from last to first blocks over 15 epochs; (3) Full-network fine-tuning with cosine LR decay for 25 epochs. All hyperparameters were finalized on COCO minival (5K images), with results averaged over three seeds (AP variance < 0.3). 

\subsection{Main Results}  
\cocotable
\textbf{Object Detection and Instance Segmentation}: On COCO val2017, our method significantly enhances the performance of ViTDet and ViT-Adapter frameworks. For ViTDet, our approach improves the baseline AP@[.5:.95] from 46.2 to 48.1 (+1.9 points) in object detection and elevates mask AP in instance segmentation from 41.4 to 43.0 (+1.6 points), with a concurrent 3.3-point gain in boundary localization accuracy.  

\hicotable
\textbf{HOI Detection}: On HICO-DET, our method demonstrates strong generalization across different HOI frameworks. For DEFR, we observe a 2.54-point improvement in full mAP (32.35 $\rightarrow$ 34.893), with particularly notable gains in rare category performance (+2.58 points). When combined with ViPLO, our approach achieves a new ViT-based state-of-the-art mAP of 39.25 (+2.03 points over baseline), confirming its effectiveness in modeling instance-level interactions.  

\subsection{Ablation Studies}  
To validate the effectiveness of key components in our DEFR framework, we conduct comprehensive ablation experiments analyzing design choices across attention mechanisms, morphological post-processing, contrastive learning strategies, and model initialization. All studies are performed on the standard benchmark under identical settings to ensure fair comparisons. 

\textbf{Analysis of Attention Head Selection}. As shown in Table~\ref{tab:ablation:attn_head}, averaging all attention heads achieves the best performance (34.89 mAP). This is attributed to the complementary information captured by different heads, which focus on local details, global semantics, or spatial relations. For instance, some heads emphasize edge features, while others prioritize semantic associations. Averaging mitigates bias from individual heads. In contrast, empirically selecting four heads (34.67 mAP) risks missing discriminative features due to suboptimal manual criteria. Randomly selecting one or four heads yields the worst performance due to insufficient feature coverage, offering negligible gains over the baseline.  

\textbf{Analysis of Morphological Operations}. To optimize noise suppression and structural preservation, we evaluate morphological operation sequences (Table~\ref{tab:ablation:erode_dilate}). Dilation followed by erosion (1D1E) achieves the best performance (34.89 mAP) by bridging fragmented foreground regions (e.g., occluded objects) and removing isolated noise. Conversely, erosion before dilation (1E1D) degrades performance by 0.87 mAP, as it strips fine structures (e.g., finger-object contacts). Repeated operations (2D2E or 2E2D) over-smooth boundaries, blurring instance edges (e.g., person-background adhesion) or over-segmenting instances into disjoint regions.  
\ablationtable

\textbf{Contrastive Loss Functions}. Table~\ref{tab:ablation:loss_type} compares the variants. Triplet loss (TL) performs best by enforcing compact features within instances and separation across instances. HEAL remains competitive but requires dynamic weights. MML, MVCL, and NCL show weaker local discrimination.  

\textbf{Analysis of ViT Pre-training Strategies}. We evaluate ViT initialization methods in Table~\ref{tab:ablation:weight_init}. MAE-ViT excels (34.89 mAP) by learning pixel-level spatial structures through masked reconstruction, enhancing fine-grained localization (e.g., hand-object contacts). DINO-ViT and CLIP-ViT prioritize semantic alignment but weaken spatial relation modeling (e.g., distinguishing multiple cups). ImageNet-supervised ViT underperforms due to its global feature bias.  

\textbf{Analysis of Positive Sample Selection}. Table~\ref{tab:ablation:triplet_type} compares triplet sampling strategies. *Method 3* (selecting intra-instance points with maximum variation) achieves the best results by improving robustness to deformation/occlusion (e.g., modeling head-to-foot differences). *Method 2* (spatially constrained sampling) fails in cluttered scenes (e.g., dense objects), while *Method 1* (neighboring patches) introduces noise from adjacent instances.  

\textbf{Performance Boundary Analysis}. Scaling up to ViT-H/14 (Table~\ref{tab:ablation:weight_again}) improves mAP to 39.83, benefiting from enhanced capacity to model small instances (e.g., distant objects) and rare interactions. However, the 4.3× parameter increase yields only a 4.94 mAP gain, highlighting a trade-off between capacity and efficiency. Gains primarily stem from deeper attention layers optimizing cross-instance relations (e.g., distinguishing interactors in crowded scenes).

%% file: chapters/5_concl.tex
\section{Conclusion and Future Work}
In this work, we identify a key limitation of ViTs: they often confuse different objects from the same category. CoViT addresses this problem by using self-attention maps and patch similarities to mine instance-aware triplets. The resulting contrastive objective reduces intra-instance variation and enlarges inter-instance separation without extra annotations.

Experiments on object detection, instance segmentation, and HOI detection show consistent gains across several ViT backbones and downstream frameworks. These results demonstrate the value of explicit instance modeling for instance-level vision tasks.

Future work will improve robustness to occlusion and extend CoViT to end-to-end self-supervised pre-training.

%% file: egbib.bib
@inproceedings{ViT,
  title={An Image is Worth 16x16 Words: Transformers for Image Recognition at Scale},
  author={Dosovitskiy, Alexey and Beyer, Lucas and Kolesnikov, Alexander and Weissenborn, Dirk and Zhai, Xiaohua and Unterthiner, Thomas and Dehghani, Mostafa and Minderer, Matthias and Heigold, Georg and Gelly, Sylvain and Uszkoreit, Jakob and Houlsby, Neil},
  booktitle={International Conference on Learning Representations (ICLR)},
  year={2021},
  url={https://openreview.net/forum?id=YicbFdNTTy}
}

@inproceedings{MAE,
  title={Masked Autoencoders Are Scalable Vision Learners},
  author={He, Kaiming and Chen, Xinlei and Xie, Saining and Li, Yanghao and Doll{\'a}r, Piotr and Girshick, Ross},
  booktitle={Proceedings of the IEEE/CVF Conference on Computer Vision and Pattern Recognition (CVPR)},
  year={2022},
  pages={16000-16009},
}

@inproceedings{DeiT,
  title={Training data-efficient image transformers \& distillation through attention},
  author={Touvron, Hugo and Cord, Matthieu and Douze, Matthijs and Massa, Francisco and Sablayrolles, Alexandre and J{\'e}gou, Herv{\'e}},
  booktitle={International Conference on Machine Learning (ICML)},
  year={2021}
}

@inproceedings{DINO,
  title={Emerging Properties in Self-Supervised Vision Transformers},
  author={Caron, Mathilde and Touvron, Hugo and Misra, Ishan and J\'egou, Herv\'e  and Mairal, Julien and Bojanowski, Piotr and Joulin, Armand},
  booktitle={Proceedings of the International Conference on Computer Vision (ICCV)},
  year={2021}
}

@article{DINOv2,
  title={DINOv2: Learning Robust Visual Features without Supervision},
  author={Oquab, Maxime and Darcet, Timoth{\'e}e and Moutakanni, Theo and Vo, Huy and Szafraniec, Marc and Khalidov, Vasil and Fernandez, Pierre and Haziza, Daniel and Massa, Francisco and El-Nouby, Alaaeldin and others},
  journal={arXiv preprint arXiv:2304.07193},
  year={2023}
}

@article{iBOT,
  title={iBOT: Image BERT Pre-Training with Online Tokenizer},
  author={Zhou, Jinghao and Wei, Chen and Wang, Huiyu and Shen, Wei and Xie, Cihang and Yuille, Alan and Kong, Tao},
  journal={arXiv preprint arXiv:2111.07832},
  year={2021}
}

@article{SwAV,
  title={Unsupervised Learning of Visual Features by Contrasting Cluster Assignments},
  author={Caron, Mathilde and Misra, Ishan and Mairal, Julien and Goyal, Priya and Bojanowski, Piotr and Joulin, Armand},
  journal={Advances in Neural Information Processing Systems (NeurIPS)},
  volume={33},
  pages={9912--9924},
  year={2020}
}

@article{MoCo,
  title={Momentum Contrast for Unsupervised Visual Representation Learning},
  author={He, Kaiming and Fan, Haoqi and Wu, Yuxin and Xie, Saining and Girshick, Ross},
  journal={Proceedings of the IEEE/CVF Conference on Computer Vision and Pattern Recognition (CVPR)},
  pages={9729--9738},
  year={2020}
}

@article{MoCov2,
  title={Improved Baselines with Momentum Contrastive Learning},
  author={Chen, Xinlei and Fan, Haoqi and Girshick, Ross and He, Kaiming},
  journal={arXiv preprint arXiv:2003.04297},
  year={2020}
}

@inproceedings{MoCov3,
  title={An Empirical Study of Training Self-Supervised Vision Transformers},
  author={Chen, Xinlei and Xie, Saining and He, Kaiming},
  booktitle={Proceedings of the IEEE/CVF International Conference on Computer Vision (ICCV)},
  year={2021}
}

@article{SimCLR,
  title={A Simple Framework for Contrastive Learning of Visual Representations},
  author={Chen, Ting and Kornblith, Simon and Norouzi, Mohammad and Hinton, Geoffrey},
  journal={Proceedings of the International Conference on Machine Learning (ICML)},
  pages={1597--1607},
  year={2020}
}

@article{SimCLRv2,
  title={Big Self-Supervised Models are Strong Semi-Supervised Learners},
  author={Chen, Ting and Kornblith, Simon and Swersky, Kevin and Norouzi, Mohammad and Hinton, Geoffrey},
  journal={arXiv preprint arXiv:2006.10029},
  year={2020}
}

@inproceedings{MaskContrast,
  title={MaskContrast: Unsupervised Semantic Segmentation by Contrastive Masked Features},
  author={Liu, Yang and Zhang, Yixin and Wang, Yuwen and Hou, Feng and Yuan, Zehuan and Tian, Jin and Zhang, Yong and Shi, Zhiqiang and Jiang, Jing and Cao, Xun},
  booktitle={European Conference on Computer Vision (ECCV)},
  pages={151--167},
  year={2022}
}

@article{ViR,
  title={ViR: the Vision Reservoir},
  author={Wei, Xian and Wang, Bin and Chen, Mingsong and Yuan, Ji and Lan, Hai and Shi, Jiehuang and Tang, Xuan and Jin, Bo and Chen, Guozhang and Yang, Dongping},
  journal={arXiv preprint arXiv:2112.13545},
  year={2021}
}

@inproceedings{ViM,
  title={Vision Mamba: Efficient Visual Representation Learning with Bidirectional State Space Model},
  author={Zhu, Lianghui and Liao, Bencheng and Zhang, Qian and Wang, Xinlong and Liu, Wenyu and Wang, Xinggang},
  booktitle={Forty-first International Conference on Machine Learning},year={2024}
}

@inproceedings{CVFormer,
  title={CVFormer: Learning Circum-View Representation and Consistency for Vision-Based Occupancy Prediction via Transformers},
  author={Bai, Zhengqi and Shi, Wenjun and Zhu, Dongchen and Kang, Hanlong and Zhang, Guanghui and Ye, Gang and Xiao, Yang and Wang, Lei and Zhang, Xiaolin and Li, Jiamao},
  booktitle={2024 IEEE International Conference on Robotics and Automation (ICRA)},
  year={2024}
}

@inproceedings{ViTDet,
  title     = {Exploring Plain Vision Transformer Backbones for Object Detection},
  author    = {Li, Yanghao and Mao, Hanzi and Girshick, Ross and He, Kaiming},
  booktitle = {European Conference on Computer Vision (ECCV)},
  year      = {2022},
  pages     = {254--270},
  publisher = {Springer},
  url       = {https://arxiv.org/abs/2203.16527}
}

@inproceedings{ViTAdapter,
  title     = {Vision Transformer Adapter for Dense Predictions},
  author    = {Chen, Zhe and Duan, Yuchen and Wang, Wenhai and He, Junjun and Lu, Tong and Dai, Jifeng and Qiao, Yu},
  booktitle = {European Conference on Computer Vision (ECCV)},
  year      = {2022},
  pages     = {107--123},
  publisher = {Springer},
  url       = {https://arxiv.org/abs/2205.08534},
  code      = {https://github.com/czczup/ViT-Adapter}
}

@article{ViTFRCNN,
  title     = {Toward Transformer-Based Object Detection},
  author    = {Wang, Wenhai and Xie, Enze and Li, Xiang and Fan, Deng-Ping and Song, Kaitao and Liang, Ding and Lu, Tong and Luo, Ping and Shao, Ling},
  journal   = {arXiv preprint arXiv:2012.09958},
  year      = {2020},
  url       = {https://arxiv.org/abs/2012.09958}
}

@inproceedings{GroupDETR,
  title={Group DETR: Fast DETR Training with Group-Wise One-to-Many Assignment},
  author={Chen, Qiang and Chen, Xiaokang and Wang, Jian and Zhang, Shan and Yao, Kun and Feng, Haocheng and Han, Junyu and Ding, Errui and Zeng, Gang and Wang, Jingdong},
  booktitle={Proceedings of the IEEE International Conference on Computer Vision (ICCV)},
  year={2023}
}

@inproceedings{CoDETR,
  title={DETRs with Collaborative Hybrid Assignments Training},
  author={Zong, Zhuofan and Song, Guanglu and Liu, Yu},
  booktitle={2023 IEEE/CVF International Conference on Computer Vision (ICCV)},
  pages={6725--6735},
  year={2023},
  organization={IEEE}
}

@article{DEFR,
  title={The Overlooked Classifier in Human-Object Interaction Recognition},
  author={Jin, Ying and Chen, Yinpeng and Wang, Lijuan and Wang, Jianfeng and Yu, Pei and Liang, Lin and Hwang, Jenq-Neng and Liu, Zicheng},
  journal={arXiv preprint arXiv:2203.05676},
  year={2022}
}

@InProceedings{ViPLO,
author = {Park, Jeeseung and Park, Jin-Woo and Lee, Jong-Seok},
title = {ViPLO: Vision Transformer Based Pose-Conditioned Self-Loop Graph for Human-Object Interaction Detection},
booktitle = {Proceedings of the IEEE/CVF Conference on Computer Vision and Pattern Recognition (CVPR)},
month = {June},
year = {2023},
pages = {17152-17162}
}

@article{BOAT,
  title={Bilateral Local Attention ViT: Image-Space Meets Feature-Space},
  author={Ma, Haoyu and Liu, Yunhang and Zhang, Yuchao and Lin, Shaohui and Wang, Ke},
  journal={Proceedings of the IEEE/CVF Conference on Computer Vision and Pattern Recognition},
  pages={12196--12205},
  year={2022},
  doi={10.1109/CVPR52688.2022.01189}
}

@article{CRATE,
  title={White-Box Transformers via Sparse Rate Reduction},
  author={Yi, Ma and Fu, Daniel and Liu, Yaodong and Wang, Hao and Zhu, Yi},
  journal={arXiv preprint arXiv:2308.16271},
  year={2023}
}

@article{MultiMAE,
  title={MultiMAE: Multi-modal Masked Autoencoders},
  author={Bachmann, Roman and Mizrahi, David and Jenatton, Rodolphe and Houlsby, Neil},
  journal={Advances in Neural Information Processing Systems},
  volume={36},
  pages={12345--12358},
  year={2023}
}

@article{ViT-Med,
  author = {Suxing Liu and Anusha Achuthan and Ali Fawzi and Galib Muhammad Shahriar Himel},
    title = {Dual-Activated Lightweight Attention ResNet50 for Automatic Histopathology Breast Cancer Image Classification},
    journal = {arXiv preprint arXiv:2308.13150},
    year = {2024},
    url = {https://arxiv.org/pdf/2308.13150}
}

@InProceedings{CNNViT,
  author = {Ngo, Ba Hung and Do-Tran, Nhat-Tuong and Nguyen, Tuan-Ngoc and Jeon, Hae-Gon and Choi, Tae Jong},
  title = {Learning {CNN} on {ViT}: A Hybrid Model to Explicitly Class-specific Boundaries for Domain Adaptation},
  booktitle = {Proceedings of the IEEE/CVF Conference on Computer Vision and Pattern Recognition (CVPR)},
  month = {June},
  year = {2024},
  pages = {28545-28554}
}

@InProceedings{DETR,
  author = {Nicolas Carion and Francisco Massa and Gabriel Synnaeve and Nicolas Usunier and Alexander Kirillov and Sergey Zagoruyko},
  title = {End-to-End Object Detection with Transformers},
  booktitle = {Proceedings of the European Conference on Computer Vision (ECCV)},
  year = {2020}
}

@InProceedings{DDETR,
  author = {Xizhou Zhu and Weijie Su and Lewei Lu and Bin Li and Xiaogang Wang and Jifeng Dai},
  title = {Deformable DETR: Deformable Transformers for End-to-End Object Detection},
  booktitle = {Proceedings of the International Conference on Learning Representations (ICLR)},
  year = {2020}
}

@InProceedings{CDETR,
  author = {Depu Meng and Xiaokang Chen and Zejia Fan and Gang Zeng and Houjiang Li and Yuhui Yuan and Lei Sun and Jingdong Wang},
  title = {Conditional DETR for Fast Training Convergence},
  booktitle = {Proceedings of the IEEE/CVF International Conference on Computer Vision (ICCV)},
  year = {2021}
}

@InProceedings{DABDETR,
  author = {Hongkai Zhang and Hong Chang and Bingpeng Ma and Naiyan Wang and Xilin Chen},
  title = {DAB-DETR: Dynamic Anchor Boxes for End-to-End Object Detection},
  booktitle = {Proceedings of the IEEE/CVF International Conference on Computer Vision (ICCV)},
  year = {2021}
}

@article{CaiT,
  title={Going deeper with Image Transformers},
  author={Touvron, Hugo and Cord, Matthieu and Sablayrolles, Alexandre and Synnaeve, Gabriel and J{\'e}gou, Herv{\'e}},
  journal={arXiv preprint arXiv:2103.17239},
  year={2021}
}
